\documentclass{article}

\usepackage{arxiv}

\usepackage[utf8]{inputenc} 
\usepackage[T1]{fontenc}    
\usepackage{hyperref}       
\usepackage{url}            
\usepackage{booktabs}       
\usepackage{amsfonts}       
\usepackage{nicefrac}       
\usepackage{microtype}      
\usepackage{lipsum}
\usepackage{graphicx}

\usepackage{xspace}
\usepackage{array}
\usepackage{xcolor}
\usepackage{subfig}
\usepackage{amsmath, amsfonts, amssymb, mathtools}
\usepackage[normalem]{ulem}
\usepackage{graphicx}
\usepackage{pifont}

\newcommand{\cmark}{\ding{51}}%
\newcommand{\xmark}{\ding{55}}%
\newcolumntype{C}[1]{>{\centering\let\newline\\\arraybackslash\hspace{0pt}}m{#1}}

\newcommand{\black}{\color{black}}
\DeclareMathOperator*{\minimize}{minimize}

\newcommand{\OF}[0]{$F_O \, (O,x,y,z)$\xspace}
\renewcommand{\vec}[1]{\boldsymbol{\mathbf{#1}}}
\renewcommand{\matrix}[1]{\boldsymbol{\mathbf{#1}}}
\newcommand{\ten}[0]{\emph{$\boldsymbol{\tau}$}\xspace}

\newcommand{\lb}[1]{\underline{#1}}
\newcommand{\ub}[1]{\overline{#1}}
\newcommand{\ud}{\mathop{}\!\mathrm{d}}

\graphicspath{ {./images/} }

\title{Stiffness-based Analytic Centre Method for Cable-Driven Parallel Robots}

\author{
 Domenico Dona'  \\
  Department of Management and Engineering (DTG), \\ University of Padua \\ Stradella S. Nicola 3, 36100 Vicenza, Italy \\
  \texttt{domenico.dona@unipd.it} \\
   \And
 Vincenzo Di Paola \\
  Dipartimento di Ingegneria Meccanica, Energetica, Gestionale e dei Trasporti,\\Universit\`{a} degli Studi di Genova,\\ Via alla Opera Pia 15 16143 Genoa, Italy. \\
  \texttt{vincenzo.dipaola@edu.unige.it} \\
\And
 Matteo Zoppi \\
  Dipartimento di Ingegneria Meccanica, Energetica, Gestionale e dei Trasporti,\\Universit\`{a} degli Studi di Genova,\\ Via alla Opera Pia 15 16143 Genoa, Italy. \\
  \texttt{matteo.zoppi@edu.unige.it} \\
  \And
 Alberto Trevisani \\
  Department of Management and Engineering (DTG), \\ University of Padua \\ Stradella S. Nicola 3, 36100 Vicenza, Italy \\
  \texttt{alberto.trevisani@unipd.it} \\
}

\begin{document}
\maketitle
\begin{abstract}
Nowadays, being fast and precise are key requirements in Robotics. This work introduces a novel methodology to tune the stiffness of Cable-Driven Parallel Robots (CDPRs) while simultaneously addressing the tension distribution problem. In particular, the approach relies on the Analytic-Centre method. Indeed, weighting the barrier functions makes natural the stiffness adaptation. The intrinsic ability to adjust the stiffness during the execution of the task enables the CDPRs to effectively meet above-mentioned requirements. The capabilities of the method are demonstrated through simulations by comparing it with the existing approach.

\end{abstract}


\section*{Nomenclature}
\noindent
\OF inertial frame\\
\noindent
$m_L$ mass of the load $\in \mathbb{R}$\\
\noindent
$\vec{u}_i$ cable direction vector $\in \mathbb{R}^3$\\
\noindent
$\ten$ cable tensions vector $\in \mathbb{R}^m$\\
\noindent
$\vec{w}_e$ external wrench vector $\in \mathbb{R}^6$\\
\noindent
$\matrix{W}$ wrench matrix $\in \mathbb{R}^{6 \times m}$\\

\section{Introduction}

Cable-Driven Parallel Robots (CDPRs) are now well-known in both the academic and industrial sphere. Because of their architecture and components they result to be moderately priced and suitable for high-speed tasks, collaborative operations and also applications that require large workspaces \cite{Bruckmannbook}.\\
\indent
Nowadays, being fast and precise is an ever-present demand in robotics. Thus, for CDPRs, the idea is leveraging the tension distribution strategy to satisfy them. The computation of the cable tensions to be applied for both normal and special operating conditions is still a topic of ongoing research~\cite{DiPaola2023Analytic,Pott2013}. Selecting cable tensions is challenging because of unilateral actuators, i.e., they can only pull. Consequently, tension limits must be incorporated either during the planning phase \cite{trevisani2013planning} or the control phase \cite{bettega2023model}.\\
\indent
One chance of solving this problem is to formulate an optimisation problem and solve it iteratively. Linear Program (LP) based methods \cite{Borgstrom2009,Shiang2000,Oh2003} are notable for their fast convergence and possible discontinuities generated between two successive solutions \cite{DiPaola2023Analytic,DiPaola2024Break}. The same can happen also with the $\infty$-norm \cite{Gosselin2011}. A natural solution consists in using a Quadratic Program (QP) \cite{Agahi2009,Taghirad2011,gouttefarde2015versatile} whose accessibility makes it popular. Other solutions were explored for example, in \cite{Hassan2008,Hassan2011} the Dykstra algorithm is used. Another alternative is taking the barycenter of the polyhedron as the optimal solution~\cite{Mikelsons2008}. A notable example is the Improved-Closed Form method \cite{Pott2013}, built on its predecessor \cite{Pott2009}, that enables real-time application. Even so, these methods do not guarantee convergence across the entire Wrench Feasible Workspace (WFW) \cite{Bouchard2010}. A strategy for tension distribution dealing with CDPRs operating beyond their WFW is also explored in~\cite{Cote2016}.\\
\indent
Moving from a global context to a particular one, where the stiffness of the system needs to be increased in an adaptive manner \cite{Reichenbach2024} depending on the task, this work proposes an additional stiffness-related use of the Analytic-Centre (AC) method \cite{DiPaola2023Analytic,DiPaolaPhD}.
The importance of stiffness in the context of CDPRs has been widely enlighted in the literature \cite{verhoeven2004analysis,nguyen2014study,kraus2016force,reichenbach2021velocity}.\\
\indent
To this end, this paper introduces an adaptive tension control technique based on the AC method. This approach allows tuning the stiffness based on the weight of the barrier functions. Observe that, because of its generality, the stiffness-adaptation represents only one use, among many, of the AC.\\
\indent
This paper is structured as follows. Section~\ref{model} recalls the main equations governing the CDPRs. Section~\ref{TDAMotivation} introduces the definition of the stiffness and the possible criteria to evaluate the performances of a TDA in terms of the stiffness. In Section~\ref{example}, a tracking task is exploited to point out the main peculiarities of the proposed approach. Finally, the results are summarized in Section~\ref{conclusion}.

\section{Modeling}\label{model}
\subsection{Dynamics}
\setcounter{figure}{-1}
\begin{figure}[tb]
  \subfloat{
	\begin{minipage}[c][1\width]{0.55\textwidth}
	   \centering
	   \includegraphics[width = 0.9\linewidth]{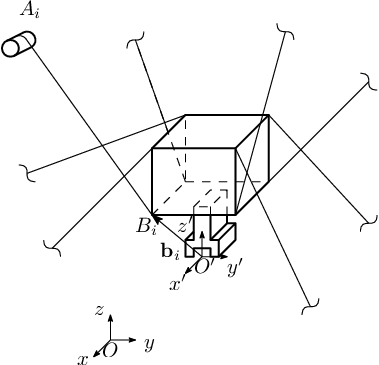}
       \caption{CDPR Scheme: general architecture of a cable-driven robot with $m$ cables and a point-mass load.}
	\end{minipage}}
 \hfill
  \subfloat{
	\begin{minipage}[c][1\width]{0.4\textwidth}
	   \centering
	   \includegraphics[scale=0.9]{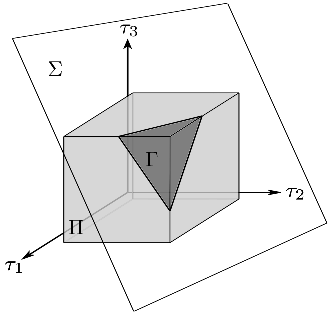}
       \caption{Feasible Solution: graphical representation of $\Pi, \Sigma$ and $\Gamma$ for the case of $m=3$. \label{fig12}}
	\end{minipage}}
\end{figure}
\setcounter{figure}{2}

In this section, the main equations and symbols necessary to describe the CDPR are hereby reported. 
The static or dynamic equilibrium of a platform in the space, guided by $m$ cables, is governed by the following equation

\begin{equation}
\matrix{W}\ten + \vec{w}_e = \vec{0}, 
\label{modelEQ}
\end{equation}

\noindent
where $\vec{w}_{e} \in \mathbb{R}^6$ is the external wrench, which also takes into account the dynamical actions applied to the platform, while $\ten \in \mathbb{R}^m$ is the cable tensions vector, and the term $\matrix{W} \in \mathbb{R}^{6 \times m}$ is the wrench matrix, which is defined as

\begin{equation}
\matrix{W} = 
\begin{pmatrix}
\vec{u}_1 & \dots & \vec{u}_m \\
\vec{b}_1 \times \vec{u}_1 & \dots & \vec{b}_m \times \vec{u}_m
\end{pmatrix},
\end{equation}

\noindent
here, $\vec{u}_i \in \mathbb{R}^3$ represents the $i$th cable direction (unitary vector) and $\vec{b}_i \in \mathbb{R}^3$ represents the $i$th attachment point on the platform. Thus, if the Degree of Redundancy\footnote{A redundant CDPR is composed of $m$ cables that exceed the Degree of Freedom (DOF) of the end-effector.} (DoR) is greater than zero, i.e. DoR$ > 0$, there exist infinite solutions of Eq.~\eqref{modelEQ} grouped in the following set

\begin{equation}
\Sigma = \big\{ \ten \, | \, \matrix{W}\ten + \vec{w}_e = \vec{0} \big\}.
\end{equation}

However, to maintain the equilibrium of the platform, the cable tension limits have to be taken into account. Hence, the $m$-dimensional convex hypercube $\Pi$ that defines the domain of the \textit{feasible} tensions is

\begin{equation}
\Pi = \big\{ \ten \, | \,\vec{0} < \lb\ten \leq \ten \leq \ub\ten \big\},
\label{tenslim}
\end{equation}

\noindent
where $\lb\ten, \ub\ten \in \mathbb{R}^{m,+}$ are positive tension vector limits; the tension limits will be considered equal to each other. Consequently, the set of feasible solutions $\Gamma$, see Fig.~\ref{fig12}, satisfying both Eq.~\eqref{modelEQ} and Eq.~\eqref{tenslim}, is

\begin{equation}
\Gamma = \Sigma \cap \Pi.
\end{equation}

\subsection{Stiffness}
Modeling the stiffness allows quantifying how precise the task would be while the robot is performing the task \cite{behzadipour2006stiffness,moradi2013stiffness}. Therefore, let us introduce the stiffness as
\begin{equation}
    \vec{K} = - \frac{\delta \vec{w}_e}{\delta \vec{q}},
\end{equation}
relating the displacement of the EE $\delta \vec{q}$ to a variation of the external wrench $\delta \vec{w}_e$. From Eq.~\eqref{modelEQ}, it follows that
\begin{equation}
    \vec{K} = \frac{\delta}{\delta \vec{q}}(\vec{W}\ten) =   \frac{\delta \vec{W}}{\delta \vec{q}} \ten + \vec{W} \frac{\delta \ten}{\delta \vec{q}} = \vec{K}_a + \vec{K}_p,
\end{equation}
where $\vec{K}_a$ is the active stiffness matrix and $\vec{K}_p$ is the passive stiffness matrix. 
The passive (or geometric) stiffness depends on the pose of the load or, in other terms, on the cable directions.
The term ``active'' underlines that can be actively tuned: indeed, the expression is the following
\begin{equation}
    \vec{K}_a = \sum_{i=1}^m \frac{1}{\ell_i} (\mathbb{I}_3 - \vec{u}_i \vec{u}_i^T) \; \tau_i,
\end{equation}
where $m$ is the number of cables, $\mathbb{I}_3$ is the matrix identity of dimension $3$, $\ell_i$ and $\tau_i$ are the length and the tension of the $i$th cable, respectively. This expression tells us that higher tensions imply higher stiffness. Therefore, it suggests considering it as a guide when formulating the tension distribution algorithms.

\section{Tension Distribution Algorithm and Comparison Metrics}\label{TDAMotivation}
\subsection{Available methods}
Typically, as discussed in Section~\ref{model}, the tension problem admits infinite solutions, necessitating the use of a cost function to select one.
Subsequently, we can select the solution that optimally complies with the chosen cost function.  \black

Several cost functions have been proposed in the literature; however, the general structure of the problem can be summarized as 
\begin{subequations}
\begin{eqnarray}
 \minimize_{\vec{\ten}} & \quad  \kappa(\vec{\ten}) \\
 \text{subject to:} & \quad \vec{W} \vec{\ten} + \vec{w}_e = \vec{0} \\ \label{eq:bounds}
 & \quad \lb\ten \leq \vec{\ten} \leq \ub\ten ,
\end{eqnarray}
\end{subequations}
where $\kappa$ is the cost function adopted. A summary of the most representative literature methods with their main features is given in Tab.~\ref{tab:criterions}. 

\begin{table}[ht!]
\centering
\caption{Literature methods comparison \cite{Pott2013,DiPaolaPhD}. \label{tab:criterions}}
\begin{tabular}{|p{3.5cm}|c|c|c|C{0.8cm}|C{1.3cm}|} 
\hline 
Method & Real-time & Workspace & Continuity & DoR & Non-Lin. Constr \\ 
\hline 
Barycentric & \textcolor{green}{\cmark} & \textcolor{green}{\cmark} & \textcolor{green}{\cmark} & 2 & \textcolor{red}{\xmark}  \\ 
Improved Closed-Form & \textcolor{green}{\cmark} & \textcolor{red}{\xmark} & \textcolor{green}{\cmark} & any & \textcolor{red}{\xmark}  \\ 
Linear Programming & \textcolor{green}{\cmark} & \textcolor{green}{\cmark} & \textcolor{red}{\xmark} & any & \textcolor{red}{\xmark}  \\ 
Quadratic Programming & \textcolor{green}{\cmark} & \textcolor{green}{\cmark} & \textcolor{green}{\cmark} & any & \textcolor{red}{\xmark} \\ 
Analytic Center & \textcolor{green}{\cmark} & \textcolor{green}{\cmark} & \textcolor{green}{\cmark} & any & \textcolor{green}{\cmark} \\ 
\hline 
\end{tabular} 
\end{table}

\begin{figure}
    \centering
    \includegraphics[width=\linewidth]{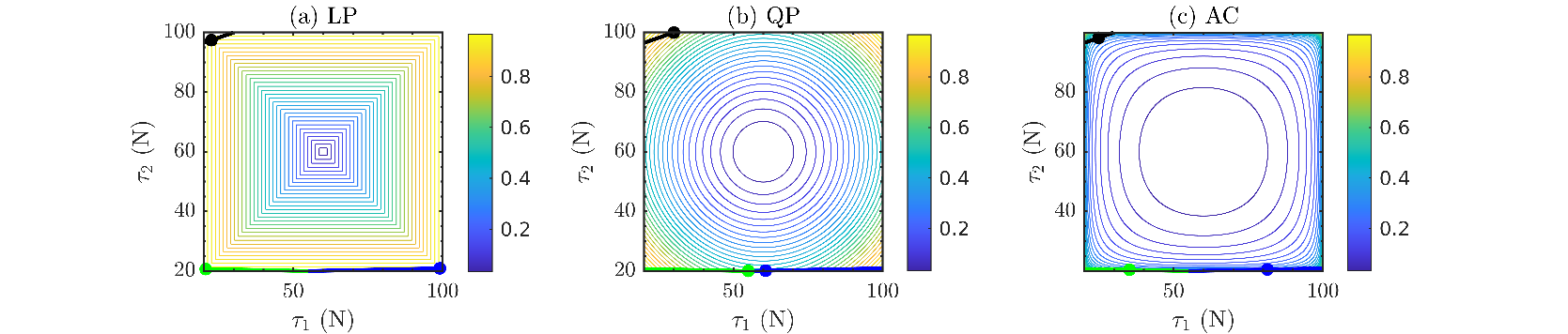}
    \caption{Level curves of Linear Programming (LP), Quadratic Programming (QP), and Analytic Center (AC). The objectives are normalized to have the same scale. Different example solutions are depicted in black, blue, and green. Refer to Table~\ref{tab:models} for the data used. }
    \label{fig:criterions}
\end{figure}

Alongside the comparison given in Tab.~\ref{tab:criterions}, a visual example is reported in Fig.~\ref{fig:criterions} for some of them. Here, a CDPR with two cables is considered with $\Pi = \{ \ten \, | \, 10 \leq  \ten  \leq 100  \}$. Three different equilibriums are depicted to simulate different robot scenarios. Among them Fig.~\ref{fig:criterions}, one can see that the Linear Programming (LP) solution exhibits the highest sensitivity: for a small variation of line angle (the blue and green lines in the picture) the tension solution suddenly steps from one corner to the other. The simulation data used are summarized in Tab.~\ref{tab:models}.

\begin{table}[ht!]
\caption{Coefficients used to compare the cost functions in Fig.~\ref{fig:criterions}.}\label{tab:models}
\centering
\begin{tabular}{|C{2.0cm}|C{2.0cm}|C{2.0cm}|C{2.0cm}|}
\hline
\# Test & Color & $\vec{W}$ [/] & $\vec{w}_e$ [N] \\
\hline
(a) & black & $[-7, 20]$ & $-1790$ \\
(b) & blue & $[-1, 50]$ & $-945$ \\
(c) & green & $[1, 50]$ & $-1055$ \\
\hline
\end{tabular}
\end{table}
When optimizing stiffness, the Adaptive Preload Control (APC) algorithm \cite{Reichenbach2024} introduces the following cost function
\begin{equation}\label{eq:ICFwS}
    \kappa (\ten) = \lVert \ten - (\eta_c \ub \ten + (1-\eta_c) \lb \ten) \rVert_2,
\end{equation}
where $\eta_c \in (0,1)$ is the so-called ``preload parameter'', that adjusts the total stiffness. The idea is to ``move'' the desired target from the mean of the bounds to a weighted mean. It is worth noting that if $\eta_c \equiv 0.5$, then the APC becomes identical to the QP algorithm \cite{Taghirad2011,Agahi2009}
\begin{equation}\label{eq:QP}
    \kappa (\ten) = \lVert \ten - \frac{1}{2}(\ub \ten + \lb \ten) \rVert_2.
\end{equation}
The limitations of the APC algorithm are the same as those of the QP, resumed in Tab.~\ref{tab:criterions}. Specifically, it cannot include nonlinear constraints and the tensions are generally not differentiable.

\begin{figure}
    \centering
    \includegraphics[width=0.8\linewidth]{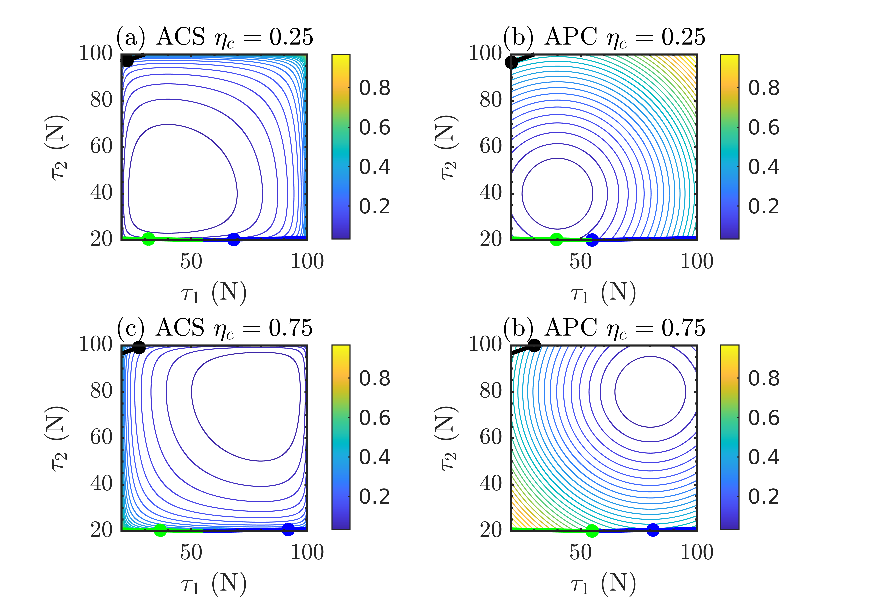}
    \caption{Level curves of Analyitc Center with Stiffness (ACS) and Adaptive Preload Control (APC) for two representative values of the preload parameter $\eta_c$. The objectives are normalized to have the same scale.}
    \label{fig:comparison}
\end{figure}

As an alternative to the QP, the AC algorithm gives a differentiable solution as well as the possibility to include nonlinear constraints. In particular, through logarithm barrier functions the inequality constraints Eq.~\eqref{eq:bounds} are enforced within the cost function
\begin{equation}
    \kappa (\ten) = - \sum_{i=1}^m \left( \log ( \tau_i - \lb  \tau)  + \log ( \ub \tau -  \tau_i) \right).
\end{equation}
The target is centered at the mean of the limit tensions, as the coefficients of the logarithms are identical, resembling the concept of the QP approach.  However, the ``pre-load'' feature of the APC can be desirable.
However, the AC method can be easily modified to include the aforementioned capabilities by using different coefficients for each term
\begin{equation}\label{eq:ACwS}
    \kappa (\ten) = - \sum_{i=1}^m \left( \lb c  \log ( \tau_i - \lb  \tau) + 
 \ub c \log ( \ub \tau -  \tau_i) \right).
\end{equation}
Uniqueness of the solution has been proved in the previous work \cite{DiPaola2023Analytic}. Defining $\alpha = \frac{\ub c}{\lb c}$, we can move the ``analytic center'' from the center of $\Pi$ to any point. Hereafter, we will refer to the Analytic Center inclusive of Stiffness as ACS. In particular, a way to connect the ACS to the APC is by taking
\begin{equation}
    \alpha = \frac{1}{\eta_c} - 1,
\end{equation}
as it can be shown that, in this scenario, the minimum of Eq.~\eqref{eq:ACwS} coincides with that of Eq.~\eqref{eq:ICFwS} in the unconstrained case. 
This new cost function allows tuning the preload by adjusting $\alpha$ (or $\eta_c$) while retaining the properties of the AC above discussed. A visual comparison between ACS and APC is reported in Fig.~\ref{fig:comparison} using the same data reported in Tab.~\ref{tab:models}.

\subsection{Comparison metrics}

To compare available TDAs and the herein proposed, we need to define some metrics.
Apart from the robustness index defined in \cite{DiPaola2023Analytic}, one could use the sensitivity to compare the TDA \cite{di2023sensitivity}. However, here we need a new metric directly related to the stiffness. Therefore, one can consider the following index
\begin{equation}
    \mathcal{I}_\textup{k} = \int_0^{t_f} \sum_{i=1}^{m} \tau_i \ud t,
\end{equation}
where $t_f$ is the total time of the task. 
Finally, we will measure the elapsed time for solving each optimization problem.

\section{Numerical Results}\label{example}

In this section, we report some numerical experiments to compare our method with the APC method. \black

We will consider two study cases, a planar CDPR with four cables and a point mass load and a spatial CDPR with a rigid body load and eight cables, depicted in Fig.~\ref{fig:4cables} and Fig.~\ref{fig:8cables}, respectively.
\setcounter{figure}{3}
\begin{figure}[tb]
  \subfloat{
	\begin{minipage}[c][1\width]{0.5\textwidth}
	   \centering
	   \includegraphics[width=\linewidth]{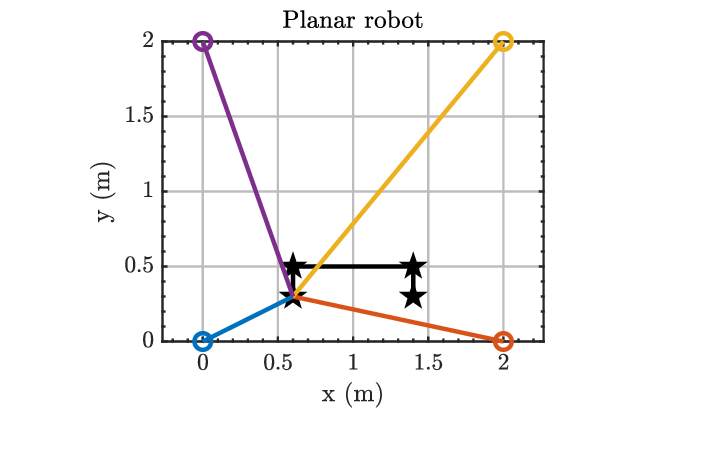}
       \caption{The four cables CDPR studied. \label{fig:4cables}}
	\end{minipage}}
 \hfill
  \subfloat{
	\begin{minipage}[c][1\width]{0.5\textwidth}
	   \centering
	   \includegraphics[width=\linewidth]{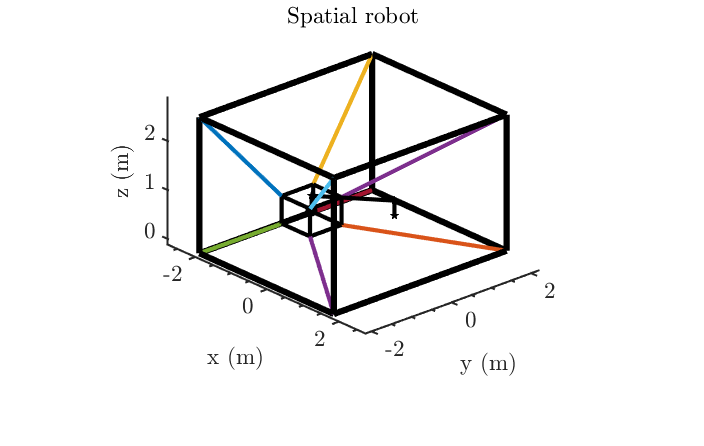}
       \caption{The eight cables CDPR studied. \label{fig:8cables}}
	\end{minipage}}
\end{figure}

For both of them, the task considered is a Pick and Place, where the stiffness is adjusted along the trajectory. The sequence begins at $P_A$, moves upwards to $P_B$ in $\Delta t_\textup{pick}$, pauses for $\Delta t_p$, then moves horizontally to $P_C$ in $\Delta t_\textup{move}$, pauses again for $\Delta t_p$, and finally moves to $P_D$ in $\Delta t_\textup{place}$. Each movement is linear and planned using a 7th-degree polynomial with zero derivatives up to the third order at the boundaries. The data used are summarized in Tab.~\ref{tab:traj}.

\begin{table}[ht!]
\caption{Points of the trajectories studied.}\label{tab:traj}
\centering
\begin{tabular}{|l|c|c|c|c|c|c|c|c|}
\hline
Robot &$P_A$ & $P_B$ & $P_C$ & $P_D$ & $\Delta t_\textup{pick}$ & $\Delta t_\textup{move}$ & $\Delta t_\textup{place}$ & $\Delta t_p$  \\
\hline
spatial & $\begin{bmatrix}
    -0.6 \\ -0.5 \\ 0.8
\end{bmatrix}$ & $\begin{bmatrix}
    -0.6 \\ -0.5 \\ 1.1
\end{bmatrix}$ & $ \begin{bmatrix}
    0.6\\ 0.5\\ 1.1
\end{bmatrix}$ & $\begin{bmatrix}
    0.6\\ 0.5\\ 0.8
\end{bmatrix}$ & $2.0$ & $5.0$ & $2.0$ & $0.5$  \\
\hline
planar & $\begin{bmatrix}
    0.6 \\ 0.3 
\end{bmatrix}$ & $\begin{bmatrix}
    0.6 \\ 0.5 
\end{bmatrix}$ & $ \begin{bmatrix}
    1.4\\ 0.5
\end{bmatrix}$ & $\begin{bmatrix}
    1.4\\ 0.3
\end{bmatrix}$ & $2.0$ & $5.0$ & $2.0$ & $0.5$  \\
\hline
\end{tabular}
\end{table}

In particular, while addressing the task we gradually change the ``pre-load parameter'' to increase the tension levels and therefore the stiffness. Again, the 7th-degree polynomials are used to transition from $0.25$ to $0.5$ during the pick, from $0.5$ to $0.75$ during the travel, and from $0.75$ to $0.9$ during the placing phase. We will use low $\eta_c$ values during the pick-phase to simulate compliance during the grasping while increasing it during the place-phase to match strict accuracy requirements. The comparison of the methods is depicted in Fig.~\ref{fig:ACwSvsAPC} and~\ref{fig:ACwSvsAPC8} for the planar and spatial robot, respectively. The sum of tension alongside the parameter $\eta_c$ is reported in Fig.~\ref{fig:SoT} and~\ref{fig:SoT8} for the planar and spatial robot, respectively.
For the spatial robot, we are only interested on the position of the end-effector, i.e. we are using a fixed orientation.
It is evident that the sum of tensions is higher in the APC than ACS. This means that the APC seems to better comply with the precision requirement. However, an important aspect for precision is also the smoothness of the tension profiles. Indeed, high variations of the tensions could induce vibration on the load which reduces precision. Hence, the combination of these two peculiarities leads us to claim that the ACS is the best compromise in terms of precision. Not to mention the possibility to include nonlinear constraints, lower computational time, and safety. Note that the APC reaches the tension limits multiple times in both study cases.

\setcounter{figure}{6}
\begin{figure}
    \centering
    \includegraphics[width=0.7\linewidth]{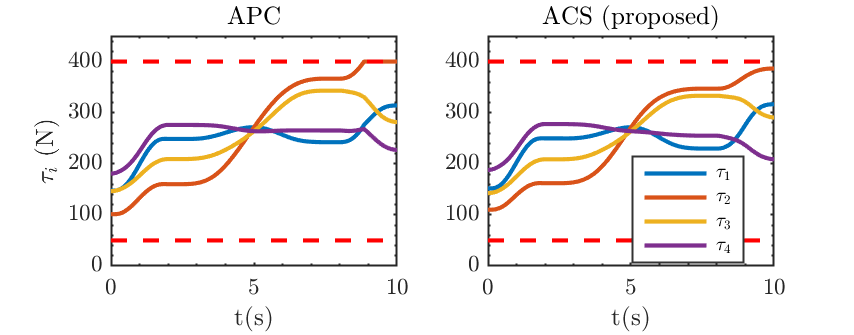}
    \caption{Cable tension results for the Adaptive Preload Control (APC) and Analytic Center with Stiffness algorithms for the planar four cables robot.}
    \label{fig:ACwSvsAPC}
\end{figure}

\begin{figure}
    \centering
    \includegraphics[width=0.7\linewidth]{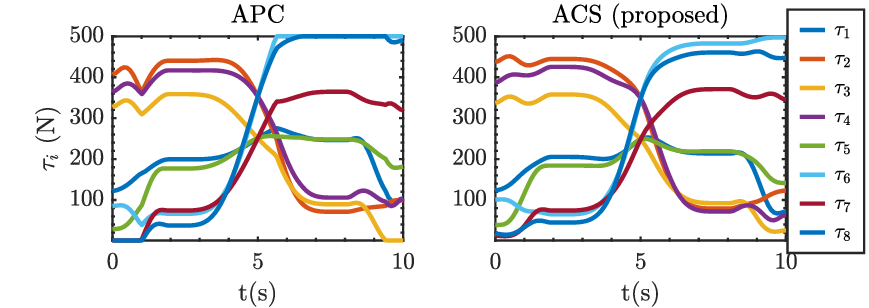}
    \caption{Cable tension results for the Adaptive Preload Control (APC) and Analytic Center with Stiffness algorithms for the spatial eight cables robot.}
    \label{fig:ACwSvsAPC8}
\end{figure}

\begin{figure}
    \centering
    \includegraphics[width=0.8\linewidth]{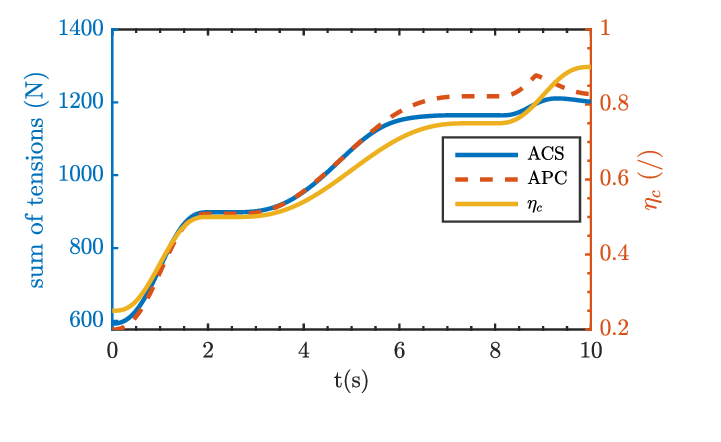}
    \caption{Sum of tensions and preload parameter evolution for the four cables robot; it is worth noting that the ACS algorithm is smoother.}
    \label{fig:SoT}
\end{figure}

\begin{figure}
    \centering
    \includegraphics[width=0.8\linewidth]{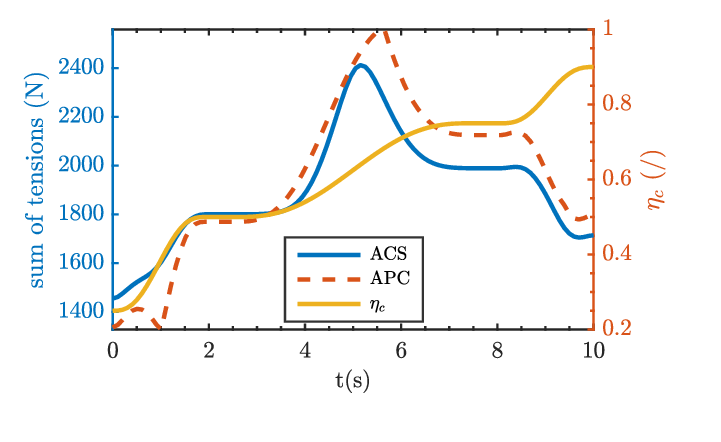}
    \caption{Sum of tensions and preload parameter evolution for the eight cables robot; it is worth noting that the ACS algorithm is smoother.}
    \label{fig:SoT8}
\end{figure}



\section{Conclusion}\label{conclusion}

In this paper, a novel method for including stiffness while computing the tension is presented.
The capabilities of the Analytic Center algorithm are shown to include also tunable stiffness, confirming its flexibility to practical applications.
The features of the proposed algorithm are demonstrated through numerical simulations. To wrap up, though the ACS does not reach the highest stiffness, it seems to perform better than the APC when speaking about precision overall without reaching tension limits and then compromising safety. From a computational point of view, it allows a higher running frequency. 

Future work aims to validate the method experimentally on physical hardware also incorporating nonlinear effects.

\section{Acknowledgment}
This study was carried out within the PNRR research activities of the consortium iNEST funded by the European Union Next-GenerationEU (Piano Nazionale di Ripresa e Resilienza (PNRR) – Missione 4 Componente 2, Investimento 1.5 – D.D. 1058 23/06/2022, ECS$\_$00000043). This manuscript reflects only the Authors’ views and opinions, neither the European Union nor the European Commission can be considered responsible for them.

\bibliographystyle{ieeetr}
\bibliography{references}

\begin{thebibliography}{10}

\bibitem{Bruckmannbook}
T.~Bruckmann and A.~Pott, ``Cable-driven parallel robots,'' {\em Springer Tracts in Advanced Robotics}, 2018.

\bibitem{DiPaola2023Analytic}
V.~Di~Paola, A.~Goldsztejn, M.~Zoppi, and S.~Caro, ``Analytic center-based tension distribution for cable-driven platforms,'' {\em Journal of Mechanisms and Robotics}, vol.~16, no.~8, 2024.

\bibitem{Pott2013}
A.~Pott, ``An improved force distribution algorithm for over-constrained cable-driven parallel robots,'' {\em 6th International Workshop on Computational Kinematics (CK)}, 2013.

\bibitem{trevisani2013planning}
A.~Trevisani, ``Planning of dynamically feasible trajectories for translational, planar, and underconstrained cable-driven robots,'' {\em Journal of Systems Science and Complexity}, vol.~26, pp.~695--717, 2013.

\bibitem{bettega2023model}
J.~Bettega, G.~Piva, D.~Richiedei, and A.~Trevisani, ``Model predictive control for path tracking in cable driven parallel robots with flexible cables: collocated vs. noncollocated control,'' {\em Multibody System Dynamics}, vol.~58, no.~1, pp.~47--81, 2023.

\bibitem{Borgstrom2009}
P.~H. Borgstrom, B.~L. Jordan, G.~S. Sukhatme, M.~A. Batalin, and W.~J. Kaiser, ``Rapid computation of optimally safe tension distributions for parallel cable-driven robots,'' {\em IEEE Transactions on Robotics}, 2009.

\bibitem{Shiang2000}
D.~C. Wei-Jung~Shiang and J.~Gorman, ``Optimal force distribution applied to a robotic crane with flexible cables,'' {\em IEEE International Conference on Robotics and Automation}, 2000.

\bibitem{Oh2003}
S.-R. Oh and S.~K. Agrawal, ``Cable-suspended planar parallel robots with redundant cables: controllers with positive cable tensions,'' {\em IEEE International Conference on Robotics and Automation}, 2003.

\bibitem{DiPaola2024Break}
V.~Di~Paola, S.~Caro, and M.~Zoppi, ``Design and performance investigation of a sliding-mode adaptive proportional--integral--derivative control for cable-breakage scenario,'' {\em Meccanica}, pp.~1--11, 2024.

\bibitem{Gosselin2011}
C.~Gosselin and M.~Grenier, ``On the determination of the force distribution in overconstrained cable-driven parallel mechanisms,'' {\em Meccanica}, vol.~46, pp.~3--15, 2011.

\bibitem{Agahi2009}
M.~Agahi and L.~Notash, ``Redundancy resolution of wire-actuated parallel manipulators,'' {\em Transactions of the Canadian Society for Mechanical Engineering}, 2009.

\bibitem{Taghirad2011}
H.~D. Taghirad and Y.~B. Bedoustani, ``An analytic-iterative redundancy resolution scheme for cable-driven redundant parallel manipulators,'' {\em IEEE Transactions on Robotics}, 2011.

\bibitem{gouttefarde2015versatile}
M.~Gouttefarde, J.~Lamaury, C.~Reichert, and T.~Bruckmann, ``A versatile tension distribution algorithm for $ n $-dof parallel robots driven by $ n+ 2$ cables,'' {\em IEEE Transactions on Robotics}, vol.~31, no.~6, pp.~1444--1457, 2015.

\bibitem{Hassan2008}
M.~Hassan and A.~Khajepour, ``Optimization of actuator forces in cable-based parallel manipulators using convex analysis,'' {\em IEEE Transactions on Robotics}, 2008.

\bibitem{Hassan2011}
M.~Hassan and A.~Khajepour, ``Analysis of bounded cable tensions in cable-actuated parallel manipulators,'' {\em IEEE Transactions on Robotics}, 2011.

\bibitem{Mikelsons2008}
L.~Mikelsons, T.~Bruckmann, M.~Hiller, and D.~Schramm, ``A real-time capable force calculation algorithm for redundant tendon-based parallel manipulators,'' {\em 2008 IEEE International Conference on Robotics and Automation}, 2008.

\bibitem{Pott2009}
A.~Pott, T.~Bruckmann, and L.~Mikelsons, ``Closed-form force distribution for parallel wire robots,'' {\em Kecskeméthy, A., Müller, A. Computational Kinematics. Springer, Berlin, Heidelberg}, 2009.

\bibitem{Bouchard2010}
S.~Bouchard, C.~Gosselin, and B.~Moore, ``On the ability of a cable-driven robot to generate a prescribed set of wrenches,'' {\em Journal of Mechanisms and Robotics}, 2010.

\bibitem{Cote2016}
A.~F. Côté, P.~Cardou, and C.~Gosselin, ``A tension distribution algorithm for cable-driven parallel robots operating beyond their wrench-feasible workspace,'' {\em 16th International Conference on Control, Automation and Systems (ICCAS)}, 2016.

\bibitem{Reichenbach2024}
T.~Reichenbach, J.~Clar, A.~Pott, and A.~Verl, ``Adaptive preload control of cable-driven parallel robots for handling task,'' {\em arXiv}, 2024.

\bibitem{DiPaolaPhD}
V.~{Di Paola}, ``Contributions to open problems on cable driven robots and persistent manifolds for the synthesis of mechanisms,'' {\em University of Genova and Ecole Centrale de Nantes, Doctoral Thesis}, 2023, \url{https://hdl.handle.net/11567/1153244}.

\bibitem{verhoeven2004analysis}
R.~Verhoeven, {\em Analysis of the workspace of tendon-based Stewart platforms}.
\newblock PhD thesis, Universit{\"a}tsbibliothek Duisburg, 2004.

\bibitem{nguyen2014study}
D.~Q. Nguyen and M.~Gouttefarde, ``Study of reconfigurable suspended cable-driven parallel robots for airplane maintenance,'' in {\em 2014 IEEE/RSJ International Conference on Intelligent Robots and Systems}, pp.~1682--1689, IEEE, 2014.

\bibitem{kraus2016force}
W.~Kraus, {\em Force control of cable-driven parallel robots}.
\newblock PhD thesis, Fraunhofer Verlag, 2016.

\bibitem{reichenbach2021velocity}
T.~Reichenbach, K.~Rausch, F.~Trautwein, A.~Pott, and A.~Verl, ``Velocity based hybrid position-force control of cable robots and experimental workspace analysis,'' in {\em International Conference on Cable-Driven Parallel Robots}, pp.~230--242, Springer, 2021.

\bibitem{behzadipour2006stiffness}
S.~Behzadipour and A.~Khajepour, ``Stiffness of cable-based parallel manipulators with application to stability analysis,'' {\em Journal of mechanical design (1990)}, 2006.

\bibitem{moradi2013stiffness}
A.~Moradi, {\em Stiffness analysis of cable-driven parallel robots}.
\newblock Queen's University (Canada), 2013.

\bibitem{di2023sensitivity}
V.~Di~Paola, S.~Caro, and M.~Zoppi, ``Sensitivity analysis of tension distribution algorithms for cable-driven parallel robots,'' in {\em International Design Engineering Technical Conferences and Computers and Information in Engineering Conference}, vol.~87363, p.~V008T08A013, American Society of Mechanical Engineers, 2023.

\end{thebibliography}
\end{document}